\documentclass[letterpaper, 10 pt, conference]{ieeeconf}  

\IEEEoverridecommandlockouts                              

\usepackage{xcolor}
\usepackage{subcaption}   
\usepackage{tabularx}
\usepackage{multirow}
\usepackage{multicol}
\usepackage{ctable}
\usepackage{dcolumn}
\usepackage{balance}
\usepackage[T1]{fontenc}
\usepackage{graphicx}
\usepackage{color, soul, cite}
\usepackage[dvipsnames]{xcolor}

\makeatletter
\let\NAT@parse\undefined
\makeatother
\usepackage[plainpages=false,hypertexnames=true,pdfnewwindow=true,backref=true,colorlinks=true,citecolor=blue,linkcolor=red,urlcolor=blue,filecolor=blue]{hyperref}

\newcommand{\fref}[1]{Fig.~\ref{#1}}
\newcommand{\sref}[1]{Sec.~\ref{#1}}

\title{\LARGE \bf

Understanding Engagement and Intrusiveness in Assistive Human-Robot Interaction Using Individual Traits
}

\author{Valerio Bo$^{1,2}$, Lavinia Hriscu$^{1,2}$, Alberto Sanfeliu$^{1,2}$ and Anais Garrell$^{1,2}$
  		\thanks{$^{1}$Institut de Robòtica i Informàtica Industrial (CSIC-UPC), 
        {\tt\small \{name.surname\}@iri.upc.edu}}
        \thanks{$^{2}$ Universitat Politècnica de Catalunya (UPC),         Barcelona, 08034, Spain.   }
\thanks{This work was supported by 
JST Moonshot R \& D grant number JPMJMS2011 and the European project TORNADO with grant number HORIZON-CL4-2024-DIGITAL-EMERGING-01-101189557. We acknowledge Fernando Herrero Cotarelo for his technical support.  
}
  \thanks{\copyright\ 2026 IEEE. Personal use of this material is permitted.
Permission from IEEE must be obtained for all other uses, in any current or future media, including reprinting/republishing this material for advertising or promotional purposes, creating new collective works, for resale or redistribution to servers or lists, or reuse of any copyrighted component of this work in other works.}}

\usepackage{graphicx}
\usepackage{amsmath}

\begin{document}

\maketitle
\thispagestyle{empty}
\pagestyle{empty}

\begin{abstract}


Robot assistance is particularly crucial in unfamiliar tasks, where users must understand task requirements while coordinating with the robot. Previous research offers mixed evidence on the role of robot proxemics in user engagement: some studies suggest closer proximity enhances interaction, while others report it can feel intrusive. In this work, we argue that perceptions of intrusiveness depend not only on proxemics but also on the frequency of robot interventions, and are strongly influenced by individual traits such as personality and demographics.
We conducted an experiment with 32 participants who interacted with two assistive robots that provided similar task support but differed in their intervention strategies. Results indicate that overall engagement remains stable across conditions, yet affective responses and perceived intrusiveness vary significantly with personality traits. Moreover, personality shapes interaction dynamics differently depending on the robot’s behavior. These findings emphasize that effective human-robot interaction should account for individual differences, tailoring robot behavior to maintain engagement while respecting each user’s unique affective and behavioral profile.
\end{abstract}

\section{Introduction} \label{sec:introduction}

Robots are increasingly deployed to assist humans in scenarios where effective support depends not only on task execution but also on accurate understanding of the user~\cite{sorrentino2022personalizing}. This need becomes critical in high-stakes domains such as healthcare, where tasks require specialized expertise and entail responsibility~\cite{zhu2025deep}. In such settings, robots need to adapt to individual users, balancing task guidance with the user's cognitive load and attention. This diversity extends beyond demographic factors to include personality traits, which deeply shape Human-Robot Interaction (HRI)~\cite{woods2007robots}.
\begin{figure}[t] 
 \centering
 \subfloat[]{%
  \label{fig:no_talk}%
  \includegraphics[width=\columnwidth]{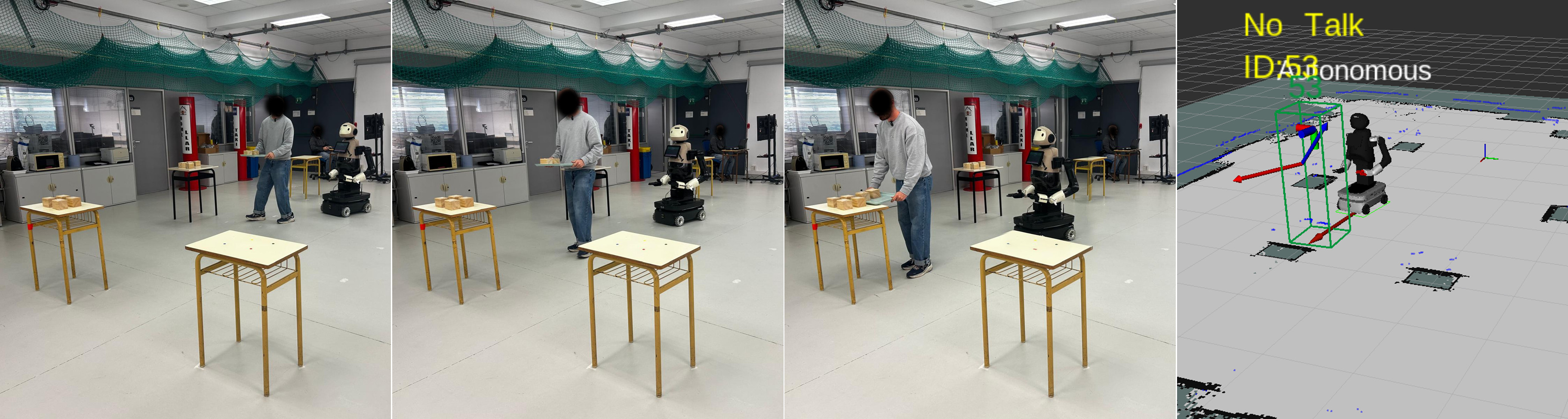}}%
 \\ \subfloat[]{%
  \label{fig:talk}%
  \includegraphics[width=\columnwidth]{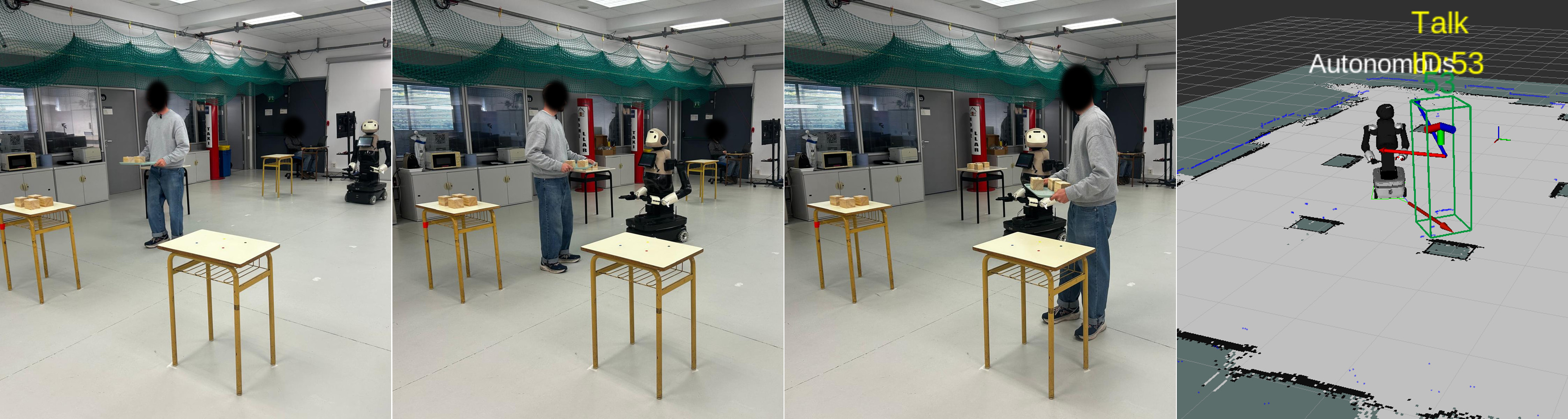}}%
 \caption{\textbf{Different robot behaviors.} (a) Sequence with a robot trying to maximize engagement by constantly following the person. (b) Sequence in which a robot approaches the person only when requested or when needed by the task.}
 \label{fig:intro}
 \vspace{-0.2cm}
\end{figure}
Engagement in this context is a dynamic balance between task goals and willingness to interact~\cite{robert2020review}. Assessing it requires observing behavioral cues while accounting for factors such as personality and demographics, which shape perceptions of robot behavior and potential intrusiveness~\cite{andriella2021have}.

While prior work has extensively explored engagement detection, less attention has been paid to how individual differences shape users' perceptions and interaction behavior. Personality traits and demographic factors influence how robot interventions are interpreted, affecting perceived intrusiveness and emotional responses~\cite{syrdal2007personalized}. At the same time, adapting robot personality to context increases engagement, highlighting the need to align behavior with user traits~\cite{andriella2021have}. However, these aspects are rarely studied jointly in task-driven scenarios, where interactions must balance user goals.

In this work, we investigate how individual differences shape both user perception and interaction behavior in HRI. We conducted a user study with 32 participants who performed a task in a simulated hospital environment, in which a robot exhibited two distinct interaction strategies, as shown in \fref{fig:intro}. We analyze how personality traits, demographic factors, and prior experience with robots relate to both subjective evaluations and objective interaction patterns. We address this research question: \textit{How do individual differences influence user perception and interaction behavior in HRI?}

The contributions of the present paper are threefold:
\begin{itemize}
\item We introduce two robot assistance intervention strategies to analyze the roles of personality traits and demographic factors in HRI. 
\item We present a multidimensional evaluation that combines subjective measures and objective interaction metrics to capture user-dependent effects.
\item We provide insights into how diversity can inform the design of personalized and adaptive robot behaviors.
\end{itemize}

The remainder of this paper is organized as follows: \sref{sec:rel_works} reviews related work, \sref{sec:method} describes the methods, \sref{sec:exp_setup} explains the experimental setup, \sref{sec:results} reports the results, \sref{sec:discussion} presents a discussion of the findings, and \sref{sec:conclusions} provides the conclusions of this study.

\section{Related Work} \label{sec:rel_works}

This Section reviews existing work on how individual differences influence human-robot interaction. We focus on how user personality shapes engagement and how robot personality and behavior affect engagement dynamics.

\subsection{Individual Differences in Human-Robot Interaction}

Existing research shows that both demographic and personality traits influence how humans perceive and interact with robots. These individual factors have been linked to perceived robot characteristics~\cite{woods2007robots} and to key interaction outcomes~\cite{esterwood2021meta, lim2022we}. Beyond robot perception, personality also shapes interaction dynamics. For instance, incorporating personality traits alongside nonverbal cues improves engagement prediction in triadic interactions~\cite{salam2017engagement} and can induce variation in engagement patterns~\cite{joshi2023pipeline}.

Personality also plays a critical role in regulating social behavior during interactions. For example, extraversion has been shown to positively influence speech duration and interaction frequency~\cite{ivaldi2017towards}. In the context of proxemics, several studies report that personality traits predict preferred interpersonal distances and expectations of robot behavior~\cite{mumm2011human}. Moreover, humans respond differently to robots than to other humans in spatial interactions, often exhibiting stronger reactions and behaviors when interacting with robots~\cite{joosse2013behave}.

Together, these findings show that individual traits not only influence human-robot interaction but also modulate how different robot behaviors are perceived and enacted, highlighting the need for interaction strategies that adapt to user-specific characteristics.

\subsection{Engagement Modeling and Adaptive Robot Behavior}

Research on engagement in HRI has largely focused on detecting and classifying engagement through observable signals, particularly visual cues such as gaze, head orientation, and body posture~\cite{khamassi2018robot}. This perspective often treats engagement as an external, measurable state, rather than a construct influenced by user-specific characteristics. More recently, multimodal approaches incorporating affective, cognitive, and behavioral signals have been proposed to provide a richer representation of engagement~\cite{rossi2021affective}.

Once engagement is estimated, robot behavior can be adapted to influence interaction outcomes. For instance, empathic responses tend to increase engagement, while more challenging or contrasting interaction styles can be perceived as stimulating~\cite{garello2020agreeableness}. Considering robot personality, extroverted robots are perceived as more useful, whereas introverted ones are considered more enjoyable~\cite{andriella2022personality}.

The perceived personality traits of humanoid robots can significantly affect users' intentions to interact~\cite{chien2022influence}. Moreover, matching extraversion or gaze behavior to the user can improve engagement, motivation, and responsiveness in rehabilitation or repetitive tasks~\cite{cruzmaya2016personality, tanevska2020adaptive}. However, these studies typically focus on single personality dimensions and settings in which the robot is the primary focus of attention.

These results highlight the importance of robot behavior design but leave open the question of how user traits mediate perception in novel task-oriented environments, or whether the same behavior might be engaging for some users and intrusive for others. To address this gap, our work investigates how individual differences shape interaction dynamics and users' perceptions when interacting with robots employing different engagement strategies in a task-driven scenario. 

\section{Methods} \label{sec:method}

This study employs two separate strategies for robot interaction, which are detailed in this Section. First, we explain the engagement module that informs both systems. We then present the designs of the two distinct architectures that use this information to govern the robot's intervention behavior. Both methods employ a real-time dialog architecture previously presented in \cite{hriscu2025human}. 

\subsection{Engagement Module}

The engagement module generates perceptual signals that inform both interaction strategies presented in this work. Specifically, it determines the participant's spatial location in the environment and extracts visual cues that reflect their orientation and attention toward the robot.

The participant's engagement level is inferred from visual cues, including body posture and head orientation. Skeleton data are obtained using Mediapipe \cite{lugaresi2019mediapipe}, a vision-based pose estimation framework that provides three-dimensional coordinates of key body landmarks. From this data, we estimate the participant's position in the environment, and derive three geometric indicators to characterize their orientation with respect to the robot:

\begin{itemize}
\item \textbf{Head orientation}: determined from facial landmarks to approximate the face normal, providing insight into whether the participant is visually attending to the robot.
\item \textbf{Body orientation}: calculated from the vector connecting the shoulders. The normal to this vector approximates the torso's forward direction, indicating whether the participant's body is directed toward the robot.
\item \textbf{Distance to the robot}: measured as the Euclidean distance between the head and the robot's base frame.
\end{itemize}

These indicators are combined to compute an engagement score, which is then used to assess the participant's engagement state. In ~\fref{fig:intro}, the RViz visualization of the predisposition to talk to the robot is shown in the last frame.

   \subsection{Rule-Based Behavior Architecture}

The first adopted system implements a reactive interaction framework that continuously receives information from the engagement module and user verbal input to guide robot interventions. The controller is structured as a finite-state machine, with the robot defaulting to a \textit{Listening} state while monitoring the participant's movements and posture, and always following the human in space, as proxemics can increase engagement~\cite{10731458}. Behavioral transitions are triggered by three event types: (i) changes in spatial position, (ii) changes in orientation relative to the robot, and (iii) verbal input detected by the speech recognition system. In this configuration, the robot actively follows the participant to remain readily available for interaction.

Changes in spatial position are treated as potential opportunities to enhance engagement, reflecting moments when the user might require assistance. Upon detecting such a change, the robot proactively offers assistance using a predefined prompt (e.g., \textit{If you need help, tell me}). If the participant responds verbally, the utterance is forwarded to the Large Language Model (LLM) for processing; otherwise, the robot returns to its passive \textit{Listening} state.
The controller tracks orientation transitions over time. If the participant turns away after previously facing the robot, a brief re-engagement prompt is issued (e.g., \textit{If you need help, tell me}). Conversely, if the participant turns toward the robot following a period of disengagement, the system waits momentarily to detect any potential requests before issuing the same prompt.
Without orientation transitions or position changes, the system checks the user's orientation whenever new verbal input is received. If the participant is not facing the robot, the system prompts them to reorient (e.g., \textit{I would appreciate it if you look at me when we are talking}). Once the participant turns toward the robot, the stored utterance is processed by the LLM, and interaction proceeds normally. 

By integrating position-based disengagement detection, attention monitoring, and verbal interaction management, this controller ensures that user requests are consistently processed whenever interaction conditions are met. The result is highly reactive, insistent robot behavior.
All predefined sentences are drawn from a set and randomly selected for the situation. For the sake of clarity and brevity, this system will be referenced as System 1.
\begin{figure}%
      \centering%
      \includegraphics[width=0.75\columnwidth]{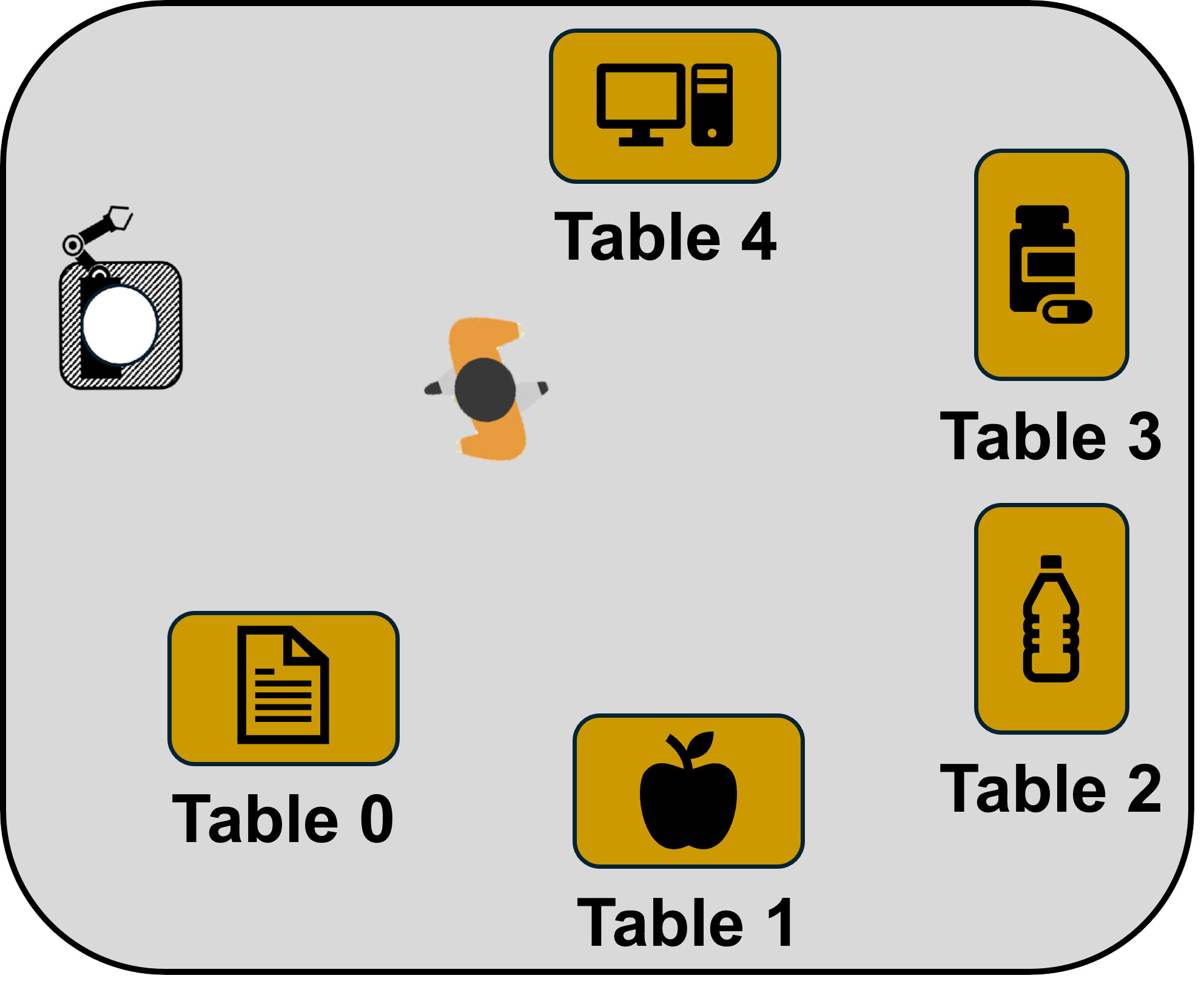}%
      \caption{\textbf{Experimental Setup.} The volunteer and the robot shared a workspace where 5 tables were present. Task instructions were placed on Table 0, food items on Table 1, fluid items on Table 2, medicines on Table 3, and the delivery station on Table 4.}%
      \label{fig:exp_setup}%
   \end{figure}%
\subsection{LLM-Driven Behavior Architecture}
The second system we tested follows a more cautious approach, relying primarily on a Large Language Model to make conversational decisions. The robot continuously tracks the participant's position, orientation, and speech activity while remaining in a default \textit{Listening} state.

In contrast to the reactive system, the LLM processes all verbal input regardless of the participant's body orientation. Thus, it interprets user requests and determines the robot's response, which may include providing a spoken reply or moving closer to assist the participant.
Whenever speech is detected, the transcribed utterance is sent to the LLM. Based on its semantic understanding, the model decides the robot's subsequent action, whether that involves delivering a verbal response or physically approaching the participant.

The system also tracks changes in the participant's orientation relative to the robot. Turning to face the robot is interpreted as a potential attempt to initiate interaction. If no user input is detected, the robot issues a short proactive prompt (e.g., \textit{I can come if you want}). The LLM then processes any verbal response to determine the appropriate action, move, or speak.

Beyond orientation cues, task context is incorporated to guide interaction. For example, when the participant reaches an area where robot intervention is needed without issuing a verbal request, the robot may offer a suggestion (e.g., \textit{I think you have to answer some unknown data. Do you want me to come and help you?}). If the participant responds, the LLM decides whether to approach or to provide verbal guidance.

All predefined sentences are randomly chosen from a set. In this system, engagement cues mainly act as triggers for proactive assistance, while the LLM handles request interpretation and response selection. Consequently, re-engagement behaviors are initiated only when contextual cues or explicit user input indicate such a need. For the sake of clarity and brevity, this system will be referenced as System 2.

\section{Experimental Setup} \label{sec:exp_setup}

In this Section, we present the experiment design, including the explanation of the task and scenario. We also provide information about the participants and the measures taken for this study. The experiments were conducted using the mobile robot platform IVO~\cite{laplaza2022ivo}, which features a camera mounted on its head. We employed Speech-to-Text (Vosk English), LLM (Qwen2.5:7B), and Text-to-Speech (Piper) models.

\subsection{Experiment Design}

\begin{figure*}
      \centering
      \includegraphics[width=0.8\textwidth]{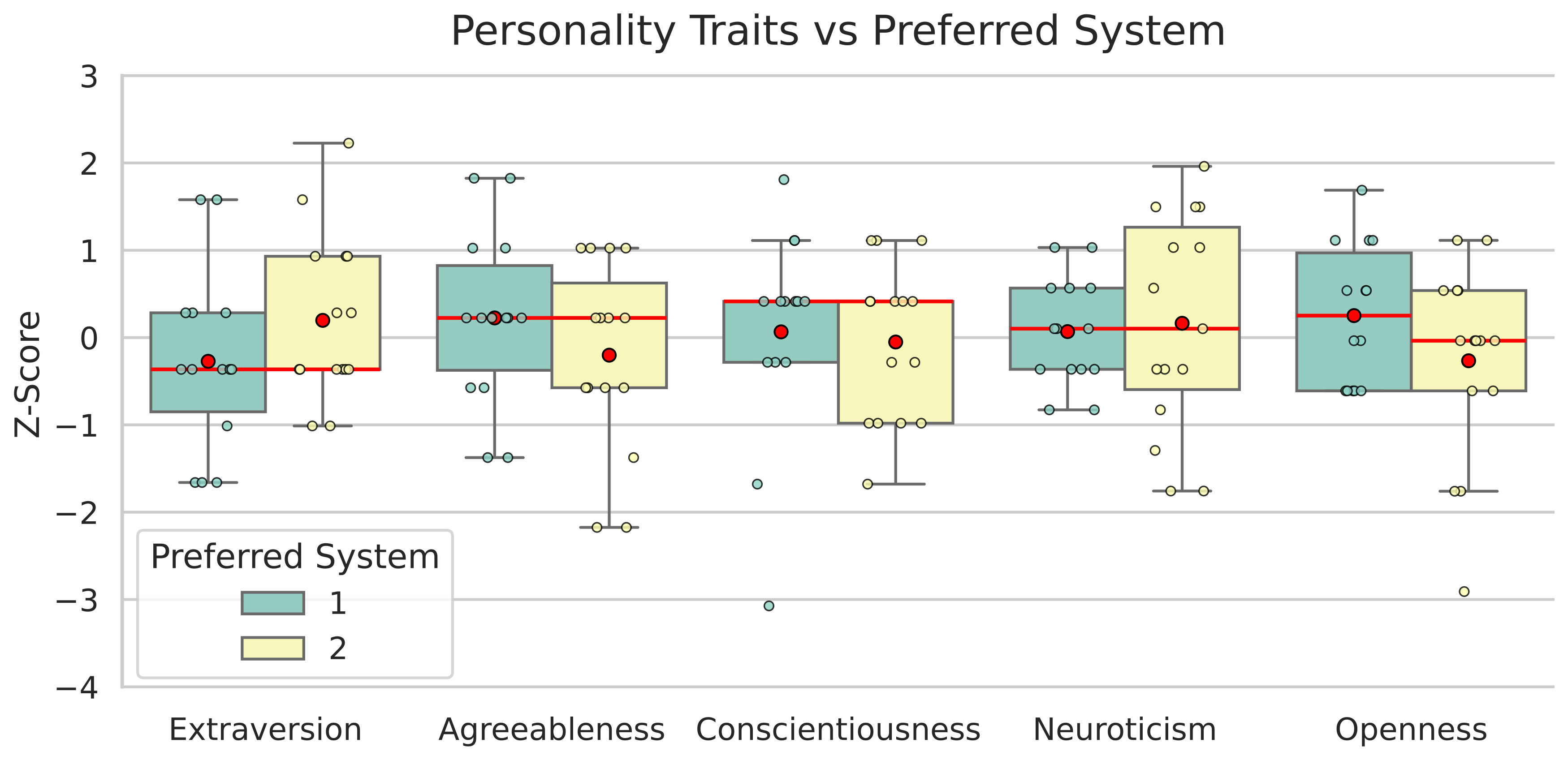}
      \caption{\textbf{Personality traits values across preferred system behavior.} Boxplots of the 5 personality traits analyzed across the preferred system for each participant. The red dots indicate the mean values and the red lines the median values. }
      \label{fig:graph_confidence}
   \end{figure*}
Participants performed a task in a simulated hospital storage unit, assuming the role of a nurse responsible for preparing items for a patient. As illustrated in \fref{fig:exp_setup}, the environment featured five tables: an initial table containing instructions and patient data, three tables with distinct categories of items, and a final table where participants placed the selected items and completed a registration form. The task was designed to require interpreting medical information in certain decisions, introducing inherent uncertainty and opportunities for robot assistance. Notably, there was no single correct combination of items, as the study did not aim to assess selection accuracy.

First, the robot approached the participant to explain the instructions and inform them that it was the sole source of assistance. In addition, participants were given instructions on paper, including a patient information sheet detailing physiological values, health conditions, and dietary restrictions, which indicated the number of items to collect from each category but did not specify which items to select. Participants were free to move among the tables and select items as instructed. Each participant completed the task twice, with the order of the two patient sheets and the robot behavior conditions randomized, and without any prior knowledge of the systems.
To ensure at least one interaction with the robot, the final step was to complete a confirmation form at the last table, which required the patient's identification code, a piece of information not available in the instructions. Participants, therefore, needed to request the code from the robot to complete the task.

A total of 32 volunteers participated in the study (19 male, 13 female), aged 19 to 65 years ($M = 31.22$, $SD = 11.18$). Before starting the experiment, participants completed a questionnaire about themselves. After each interaction, they answered questions regarding their experience and indicated their preferred system. Interactions were in English and had no time limit.
All participants provided written informed consent before the experiment using a form approved by the Universitat
Politècnica de Catalunya (UPC) under Application No. 2026.054.

\subsection{Adopted Metrics}

Before the experiment, participants completed a demographic questionnaire capturing age, gender, and prior experience with robots. Personality traits were assessed using the 10-item short version of the Big Five Inventory \cite{rammstedt2007measuring}.

Following each interaction, participants evaluated their experience using multiple measures: Godspeed questionnaire for perceived intelligence \cite{bartneck2009measurement}; Self-Assessment Manikin (SAM) for emotional response\cite{bradley1994measuring}; user engagement with the User Engagement Scale (UES) \cite{o2018practical}; and perceived intrusiveness rated on a 5-point Likert scale with items specifically adapted for this study:

\begin{enumerate}
\item The robot invaded my personal space.
\item The robot interrupted me unnecessarily.
\item The robot respected my autonomy (reverse-coded).
\item The robot intervened at inappropriate moments.
\end{enumerate}

After completing both conditions, participants indicated their preferred interaction style, based on the robot's movement and speech.
In addition to self-reported measures, interaction logs were collected, including textual conversations, interaction durations, counts of user inputs and robot responses, and the number of participant position changes.

\subsection{Hypotheses}
Guided by the research question on how individual differences influence user perception and interaction behavior in HRI, and informed by prior work on engagement and personalized robot behavior~\cite{andriella2021have}, we formulated three hypotheses linking participant demographics, personality traits, and prior experience to both subjective and objective measures of interaction with a robot that exhibited distinct engagement strategies. The hypotheses are as follows:
\begin{itemize}
    \item H1 - Participant demographics and prior exposure to robots influence system preference.
    \item H2 - Engagement shows limited variation across personality traits, while individual differences could still modulate perceived experience.
    \item H3 - Personality traits relate to both subjective evaluations and objective measures of interaction.

\end{itemize}

Together, these hypotheses examine how demographic and personality factors interact with robot behavior to shape user perceptions and interaction dynamics in HRI.

\section{Results} \label{sec:results}
In this Section, we present the results obtained from the analysis of participant and interaction data. It should be noted that no system was preferred by the majority, since System 1 received 14 votes and System 2 received 15. Three volunteers reported no clear preference between the two systems, and we removed those records when analyzing correlations between individual traits and preferred system.

\subsection{Participant Demographics}
First, to examine the relationship between demographic factors (age, gender, prior robot interaction) and preferred robot system (System 1 vs. System 2), we conducted a logistic regression, including the system used in the first experiment, to account for potential bias due to exposure order. Age was not a statistically significant predictor of system preference; however, a descriptive marginal trend suggested that participants over 50 years old tended to prefer System 2, potentially reflecting a preference for less intrusive, more contextually adaptive interactions. Gender, in contrast, emerged as a significant predictor: males were significantly less likely to prefer System 2 than females ($\beta=-2.472$, $p=0.041$, $95\%$ $CI = [-4.846, -0.099]$), suggesting a greater tolerance for the highly reactive behavior characteristic of System 1. Regarding prior experience with robots, results were mixed and inconclusive. Participants with daily or occasional exposure to robots tended to prefer System 1. Conversely, those with multiple past interactions showed a slight tendency toward System 2, though this subgroup was small and the trend did not reach statistical significance. Finally, no significant effect of interaction order was observed, suggesting that novelty or fatigue effects did not systematically bias system preference.

\subsection{Personality Traits}
\paragraph{System Preference} We also investigated whether personality traits were related to system preferences using logistic regression. As shown in \fref{fig:graph_confidence}, Extraversion exhibited a weak, non-significant tendency toward System 2 ($p = 0.145$, $95\%$ $CI = [-0.265, 1.799]$), which may reflect a preference for its more flexible interaction style over the more rigid, rule-based prompts of System 1. In contrast, Openness showed a marginal negative association with preferring System 2 ($\beta = -1.017$, $p = 0.072$, $95\%$ $CI = [-2.124, 0.090]$). No meaningful effects were observed for Agreeableness, Conscientiousness, or Neuroticism.

The test revealed comparable average engagement scores for the two systems (System 1: $3.64$ with $SD = 0.60$; System 2: $3.57$ with $SD = 0.69$). Neither personality traits nor the order of exposure significantly influenced engagement.

Analysis of the SAM questionnaire revealed a marginal negative association between Openness and Arousal ($\beta = -0.745$, $95\%$ $CI = [-1.510, 0.020]$, $p = 0.056$). Valence showed marginal significant association with being extraverted ($\beta = 0.526$, $95\%$ $CI = [-0.028, 1.080]$, $p = 0.063$). No effects were found for Dominance.

Perceived Intelligence showed no correlation with personality traits, while Perceived Intrusiveness differed significantly between the two architectures. System 2 was repeatedly rated as less intrusive than System 1 ($\beta = -0.977$, $95\%$ $CI = [-1.364, -0.589]$, $p < 0.001$), consistent with its more selective intervention strategy, which relies on contextual cues or explicit input rather than frequent re-engagement prompts. Neuroticism was positively associated with Perceived Intrusiveness ($\beta = 0.289$, $95\%$ $CI = [0.037, 0.541]$, $p = 0.025$).

\begin{figure}%
    \centering%
    \includegraphics[width=0.89\columnwidth]{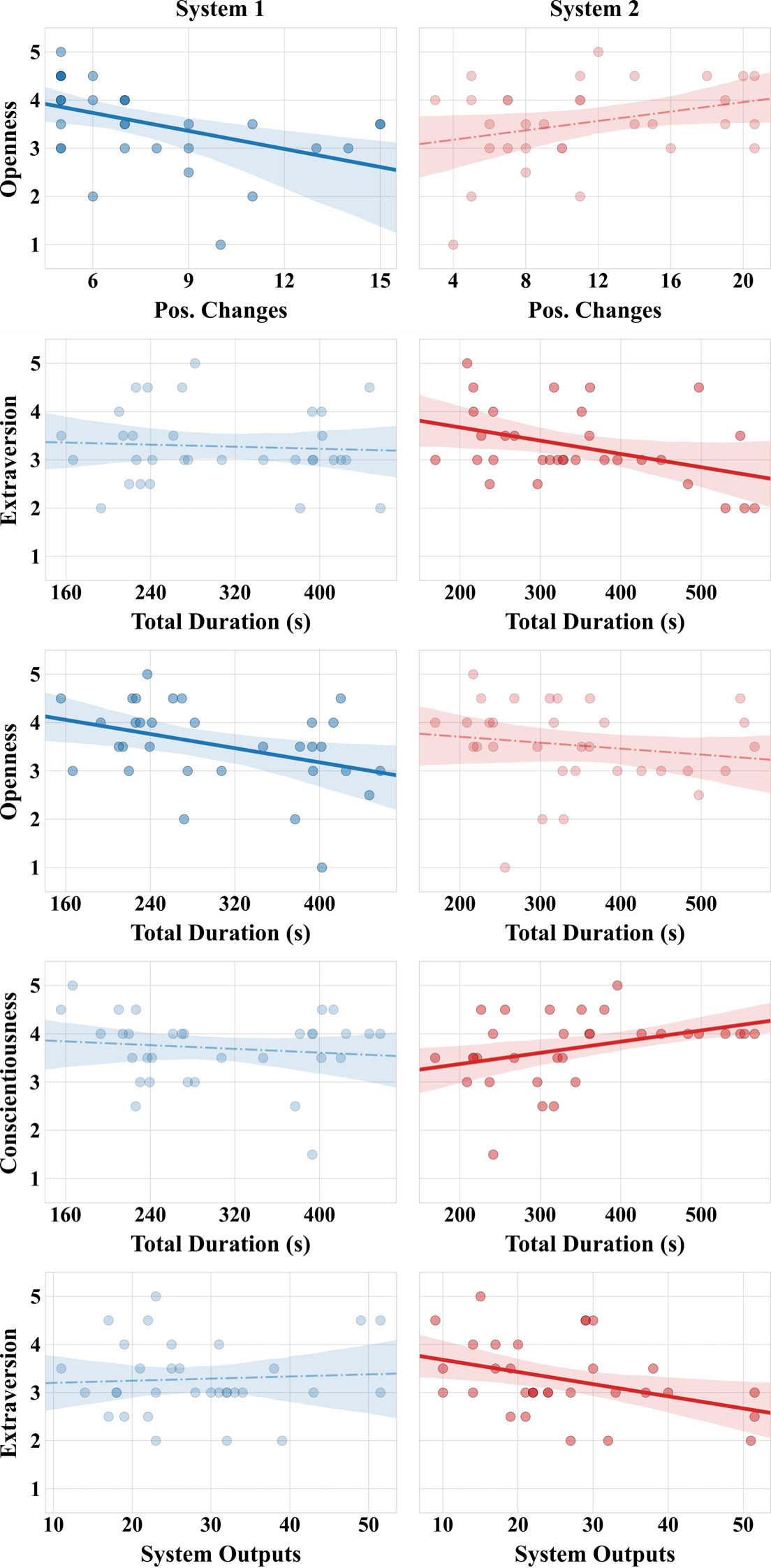}%
    \caption{\textbf{Correlations between objective and personality metrics.} Scatter plots showcasing correlations across both systems. Solid lines indicate significance with $p < 0.05$; dashed lines indicate $p \geq 0.05$.}%
    \label{fig:correlation_obj_subj}%
\end{figure}

\paragraph{Objective Metrics} Objective interaction metrics were further analyzed to examine relationships between interaction and personality traits. Spearman correlations, applied due to non-normality, revealed distinct patterns (\fref{fig:correlation_obj_subj}). In System 1, Openness was negatively correlated with interaction duration and movement, suggesting more efficient task completion and a tendency to assimilate the proactive cues quickly. In System 2, Extraversion was negatively correlated with both interaction duration and the number of robot outputs, indicating shorter, more user-driven interactions, likely due to participants taking conversational initiative. Conversely, Conscientiousness was positively correlated with interaction duration in System 2, reflecting a more methodical and thorough approach to task engagement. No other correlations reached significance.

\section{Discussion} \label{sec:discussion}

This Section discusses the results obtained, also taking into account the open comments provided by participants, and relates all findings to wider evidence.

\subsection{Individual Traits Effects on Subjectivity
}

While system preferences were broadly balanced, more nuanced patterns emerged when considering individual characteristics, suggesting that user experience is shaped by the interplay between system behavior and personal factors.

In accordance with prior findings that demographics can shape perceptions of robot behavior~\cite{woods2007robots}, we found gender to be the only significant predictor in our study, with female participants showing a stronger tendency to prefer System 2. This trend posits a more implicit mechanism, raising the possibility that the gender effect reflects differences in tolerance for interaction styles, such as responsiveness to interruptions or control asymmetry, rather than a direct preference for specific robot behaviors. Age and prior exposure had no significant effect, suggesting that demographic factors are less impactful than personal traits. Collectively, these results provide partial support for H1, indicating that demographic influences are present but limited.

Personality traits offered additional insights, though effects were generally subtle. Extraverted profiles showed a preference for non-invasive systems, whereas participants higher in Openness did the opposite, suggesting that greater curiosity and openness to experience may lead to a preference for systems that promote continuous interaction, even when it's not necessary for task success.  Traits including Agreeableness, Conscientiousness, and Neuroticism did not predict system preference, suggesting that preference formation is more closely linked to exploration- and social-engagement-related dimensions than to compliance or emotional stability. 

Participants were largely unfamiliar with the hospital context. They reported feeling immersed in the task and described the experience as engaging and interesting, which may explain the similar engagement scores observed across systems, regardless of personality effects. In this setting, engagement appears to reflect the overall experience rather than being driven solely by the system's interaction strategy.

In contrast, emotional response was related to personality: more curious participants, who were more prone to experience greater novelty, felt less excited and activated, potentially reflecting faster cognitive adaptation to the interaction. Furthermore, more extroverted participants showed a marginal tendency to be more pleased with the less interventionist system, which may be related to feeling more relaxed and unconstrained, as some stated in the open comments.

A clearer distinction between systems emerged in this analysis. System 2 was designed to minimize unnecessary interventions and to act primarily when contextual cues or explicit input indicated a need, thereby reducing intrusion. Results indicate that the effect was indeed perceived by participants, who were sensitive not only to the frequency of robot actions but also to their necessity and timing, especially among users with more negative, stressed attitudes. Together, these results support H2 and highlight perceived intrusiveness as a key dimension through which both system design and individual differences shape user experience.
\subsection{Personality Impact on Interactions}

The analysis of objective interaction metrics indicates that personality effects are not uniform but depend on the robot's interaction strategy. Rather than acting as stable predictors, personality traits are amplified or attenuated depending on the level of control the system exerts.

In System 1, which enforced a more structured interaction, the effects of personality on interaction dynamics were limited. The only significant relationship was between Openness and both interaction duration and user movement, with Openness negatively correlated with each. This result suggests that more open and imaginative individuals adapted quickly to the system's behavior, requiring less exploration and completing the task more efficiently. The structured interaction likely constrained behavioral variability in user actions, reducing the expression of other personality differences.
In contrast, for System 2, extraverted users engaged in shorter, less robot-driven interactions, likely taking a more active role in guiding the task. The opposite was observed among participants who were more perfectionistic and conscientious, suggesting that when given greater autonomy, they spent more time on the task because they were more thorough and methodical.

These contrasting patterns highlight a key insight: the robot's interaction strategy not only shapes user behavior but also determines the extent to which individual differences are expressed. Structured interactions dampen personality effects, whereas flexible interactions allow them to emerge. These results support H3 and suggest that personality's impact on HRI is conditional on the distribution of control between the human and the robot.

\section{Conclusion} \label{sec:conclusions}
This work investigated how individual differences shape user perceptions and behaviors in task-based Human-Robot Interaction through a controlled study with 32 participants in a simulated hospital environment, comparing two interaction strategies using both subjective and objective measures.

Results indicate that while task demands primarily drove engagement, individual traits significantly shaped perceptions of intrusiveness, affective responses, and interaction patterns. Neuroticism heightened Perceived Intrusiveness, whereas Openness influenced Arousal, preferences, and Efficiency. Also, Gender predicted system choice. Generally, personality-driven differences in interaction dynamics, such as interaction duration and communication patterns, depend on system design: adaptive strategies accommodate heterogeneity, whereas reactive systems constrain it. Personalization must address not only robot actions but also the level of shared control, as this directly shapes how heterogeneity emerges during interaction.

Future work will examine how engagement, trust, and collaboration evolve over repeated interactions, enabling robots to co-adapt with users over time. Additionally, research will extend to multi-step collaborative tasks and multi-user scenarios \cite{bo2026fast}, investigating how individual differences shape coordination, role negotiation, and group dynamics in more complex social and operational contexts. Given our moderate sample size, we plan to extend these experiments to further test and generalize these findings across other HRI domains.





\bibliographystyle{IEEEtran}
\bibliography{references}

\end{document}